\documentclass{article}
\usepackage{spconf,amsmath,graphicx,multirow,booktabs,array}

\makeatletter
\renewcommand{\maketag@@@}[1]{\hbox{\m@th\normalsize\normalfont#1}}%
\makeatother

\title{A  feature refinement module for light-weight semantic segmentation network}
\name{Zhiyan Wang, Xin Guo,  Song Wang*, Peixiao Zheng, Lin Qi \thanks{*Corresponding author: Song Wang, E-mail: ieswang@zzu.edu.cn. This research was supported by the National Natural Science
Foundation of China under Grant No.62101503}}
 
\address{ School of Electrical and Information Engineering, Zhengzhou University,  China}
\begin{document}
%
\maketitle
\begin{abstract}
Low computational complexity and high segmentation accuracy are both essential to the real-world semantic segmentation tasks. However, to speed up the model inference, most existing approaches tend to design light-weight networks with a very limited number of parameters, leading to a considerable degradation in accuracy due to the decrease of the representation ability of the networks. To solve the problem, this paper proposes a novel semantic segmentation method to improve the capacity of obtaining semantic information for the light-weight network. Specifically, a feature refinement module (FRM) is proposed to extract semantics from multi-stage feature maps generated by the backbone and capture non-local contextual information by utilizing a transformer block. On Cityscapes and Bdd100K datasets, the experimental results demonstrate that the proposed method achieves a promising trade-off between accuracy and computational cost, especially for Cityscapes test set where 80.4\% mIoU is achieved and only 214.82 GFLOPs are required.
\end{abstract}

\section{Introduction}
\label{sec:intro}
Semantic segmentation aims to assign semantic labels for every pixel in the image, which has been applied in various computer vision applications, like intelligent driving \cite{Driving}, robot navigation \cite{Robots}, medical image analysis \cite{Medical} and so on. Since Long et al. \cite{FCN} firstly introduced full convolution network into the segmentation task, semantic segmentation has achieved great performance gains. However, the performance gains heavily rely on deep networks that hardly meet the speed demand of model inference in practical applications due to the expensive computation cost.

To alleviate this problem, plenty of light-weight backbones \cite{lightweight1,lightweight2,lightweight3,lightweight4}  with shallow structure and few parameters are introduced into segmentation networks. However, the semantic extraction ability of those light-weight backbones is weaker than that of deep ones, leading to the degradation of segmentation accuracy.
In this case, some researches proposed to explore global context information from the extracted semantic feature maps to improve the accuracy of label assignment for pixels. In ParseNet \cite{ParseNet}, Global average pooling was employed to extract global context by pooling the feature from the last stage of the backbone. Inserting the global features into the local ones, ParseNet greatly improved the quality of semantic segmentation. Pyramid pooling module \cite{PPM} was proposed to fuse features under different pyramid scales to reduce the context loss between sub-regions. Based on PPM, Deep Aggregation Pyramid Pooling Module (DAPPM) \cite{DDRNet} extracted context information under more pyramid scales, and proposed a more complex fusion strategy for the multi-scale contexts. However, there are some drawbacks about the above methods. (1) In those above pooling strategies, the contexts are captured by averaging all pixels in the pooling region, which neglects that different pixels in the pooling region may make unequal contributions to the semantic label assignment for a given pixel. The pixels with same labels usually yield more useful information than those with different labels. (2) Those above methods focus on the semantic information of the feature map from the last stage of the backbone, ignoring the semantics from other stages. Actually, semantic information can be obtained from the feature maps of all the stages.

 To overcome these drawbacks, we propose a novel light-weight semantic segmentation networks for accurate and fast segmentation. Specifically, a feature refinement module (FRM) is designed to aggregate semantic information from all four stages in the backbone by pooling the multi-stage features to the same size and concatenating them together. In addition, the attention operation in FRM explores the correlation between pixels and captures reasonable context information. The contributions of this paper are summarized as follows:
 
(1) A novel semantic segmentation algorithm is proposed to improve the ability of light-weight network for accurate and fast semantic segmentation. 

(2) The feature refinement module is designed to extract rich semantic information by aggregating multi-stage feature maps and capture reasonable contexts with the utilization of the attention mechanism.

(3) The proposed method achieves a better trade-off between accuracy and computational cost on the benchmarks of Cityscapes and Bdd100K than other SOTA algorithms. The experimental results of our method on the Cityscapes test set reach to 80.4\% mIoU with 214.82 GFLOPs for a $2048\times1024$ input.

\section{The proposed method }
\label{sec:majhead}
This section describes the details of the proposed method for segmentation. We first present the whole framework of the proposed method. Then, the feature refinement module which extracts and refines semantic information will be introduced in Sec. 2.2. After that, Sec. 2.3 displays the hybrid loss function used in the proposed method.

\subsection{Network Architecture}
\label{ssec:subhead}
As shown in Fig. 1, the proposed light-weight semantic segmentation network is based on the encoder-decoder structure. The encoder includes the backbone and the feature refinement module. There are four stages in the backbone, where the output of each stage is denoted as $\left\{F_1,F_2,F_3,F_4\right\}$. The output size of each stage is $\{\frac{H}{4}\times\frac{W}{4},\frac{H}{8}\times\frac{W}{8},\frac{H}{16}\times\frac{W}{16},\frac{H}{32}\times\frac{W}{32}\}$, where $H$ and $W$ indicate the height and the width of the input image respectively. FRM extracts semantic information from the multi-stage features and outputs the semantic features. The details of FRM based on a transform block are elaborated in Sec. 2.2. The FPN architecture is used as the decoder following the setting in \cite{SFNet}. 

\subsection{Feature Refinement Module}
\label{ssec:frm}
To extract rich semantics and refine semantic features for accurate and fast semantic segmentation, a feature refinement module (FRM) is proposed in this section, as shown in Fig. 2. Instead of extract semantics only from the last stage of the light-weight backbone, FRM aggregates all the four stages feature to take full advantage of the semantic information from all the stages of the light-weight backbone. Specifically, FRM firstly pools the multi-stage features with different scales to the same size  $\frac{H}{32}\times\frac{W}{32}$ and concatenates them together. The concatenated feature $F_c$ is computed as 
\begin{equation} 
\!F_c\!=\!cat(pooling(F_1),pooling(F_2),pooling(F_3),F_4),
\end{equation}
where $cat()$ and $pooling()$ denote the concatenate operator and the average pooling operator.

To capture global contextual information, the disentangled non-local block (DNL) \cite{DNL} is employed to explore the correlation between different positions in the concatenated feature map. By evaluating the similarity between pixels, DNL block adaptively weights all the positions to capture the contribution-dependent contexts. Specifically, the context for each pixel is evaluated by computing the weighted sum of all pixels in the concatenated feature map $F_c$. With $x_i$ representing the value at position $i$, the output $y_i$ of the DNL is computed as 
\begin{equation} 
y_i=\sum_{j\in\mathrm{\Omega}}{w(x_i,x_j)g(x_j)}, 
\end{equation}
where $w\left(x_i,x_j\right)$ stands for the similarity between $x_i$ and $x_j$; $g\left(x_j\right)$ represents the unary transformation for $x_j$. $\mathrm{\Omega}$ is the set of all pixels. The similarity function $w\left(x_i,x_j\right)$ is defined as 
\begin{equation} 
w\left(x_i,x_j\right)=\sigma\left({(q_i-\mu_q)}^T(k_j-\mu_k)\right)+\sigma(m_j),
\end{equation}
where $\sigma(\cdot)$ is the SoftMax function. The embeddings $q_i$, $k_j$, $m_j$ are computed as $W_qx_i$, $W_kx_j$ and $W_mx_j$ separately, where $W_q$, $W_k$\ and $W_m$ are the weight matrixes to be learned. The average of $q_i$ and $k_j$ over all pixels are denoted as  $\mu_q$\ and $\mu_k$\ respectively. The DNL block is shown in Fig. 2. Specifically, the transformation functions for $g\left(x\right)$ , $q$ , $k$, $m$ are implemented with a 1x1 convolution operator.

The feed-forward network (FFN) is utilized to further enhance the representation ability of the network. The structure of FFN includes one 1x1 convolution to expand the dimension of feature channel, a depth-wise convolution, a ReLU layer, and a 1x1 convolution to reduces the dimension of feature channel to the original number. To reduce the computational cost, a 1x1 convolution is added after FFN for cutting the channels. 

\begin{figure} 

\includegraphics[width= \linewidth] {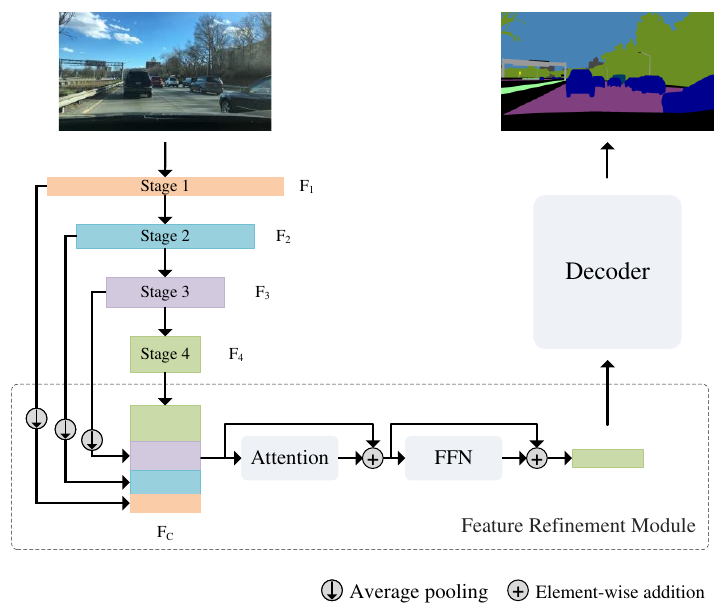}
 
\caption{The structure of our proposed approach.}
 
\end{figure} 

 \begin{figure*}[htbp]
\centering
\includegraphics[width=\linewidth]{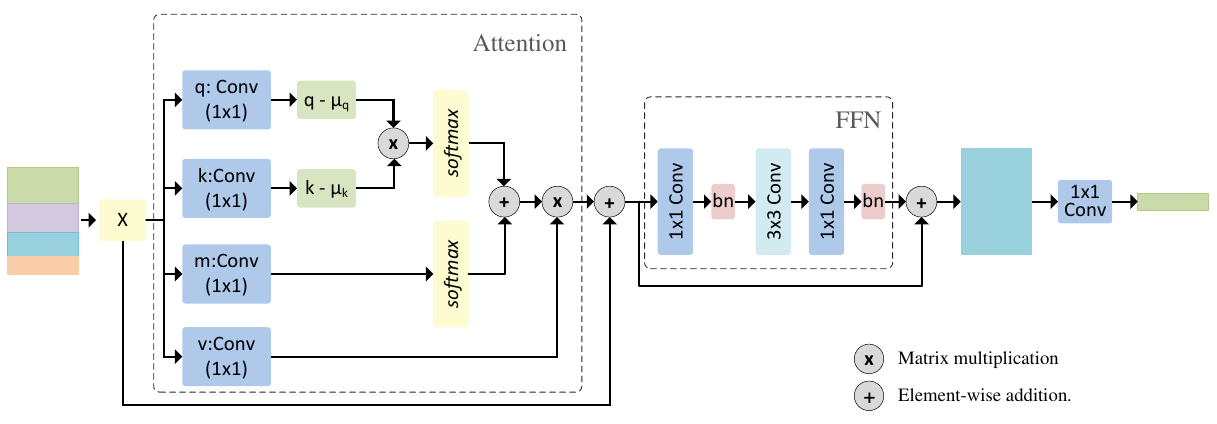}
 
\caption{ Illustration of Feature Refinement Module.}
 
\end{figure*} 
In this case, the capacity of obtaining semantic information for the light-weight networks is improved by FRM. On the one hand, FRM aggregates semantic information from multi-stage features in the backbone, which makes full use of semantics generated by the light-weight backbone. On the other hand, FRM adaptively weights the related regions to capture more reasonable contextual information. The refined semantic features from FRM is able to drastically improve the accuracy of semantic segmentation, which is further demonstrated in the Section 3.
   
\subsection{Loss Function}
\label{ssec:subhead-loss}
In this paper, a hybrid loss function is employed to train the proposed segmentation network.
The hybrid loss function is defined as 
\begin{equation}
 L=L^{ce}+\lambda\ L^{cl},
\end{equation}
where $\lambda$ is the coefficient. The cross-entropy loss $L^{ce}$ is a commonly used loss function in semantic segmentation tasks. It is defined as 
\begin{equation}
L^{ce}=-\sum_{i=1}^{n}{y_ilog{(p}_i)},
\end{equation} 
where $y_i$ denotes the ground truth and $p_i$ is the predicted probability for the given pixel $i$.

The contrastive loss \cite{closs} is introduced to learn the intrinsic structures of training data to improve segmentation accuracy. The aim of the contrastive loss is to enforce the pixels of the same class closer and the pixels of different classes far away from each other in the embedding space. The contrastive loss $ L^{cl}$ is formulated as 
\begin{equation}
\!\!\!L_i^{cl}\!=\!\frac{1}{\left|p_i\right|}\ \!\!\!\! \sum_{i^+\in P_i}\!\!\!{-log\ \!\!\frac{\exp(i\cdot i^+/\tau)}{\exp(i\cdot i^+/\tau)\!+\!\!\sum_{i^-\in \! N_i}\!\!{\exp(i\cdot i^-/\tau)}}}\!,
\end{equation}
where $\tau>0$ is temperature hyper-parameter. For the pixel $i$, $i^+$ and $i^-$ denote embeddings of samples from same class and different class respectively. To map pixel features into the embedding space, the embedding head is implemented as one 1x1 convolution layers. The embedding head is only applied during training, which means that there is no extra computational cost at inference time.

\section{Experiments}
\label{sec:Experiments}
\subsection{Datasets}
\label{ssec:Datasets} 
The performance of the proposed method is evaluated on two semantic segmentation datasets, Cityscapes \cite{Cityscapes} and Bdd100K \cite{BDD100K}. Cityscapes is an urban street scene dataset with high-resolution ($2048 \times 1024$) images. It contains 5000 fine-annotated images, of which 2975 images for training, 500 images for validation, and 1525 images for testing. Bdd100K is a new and challenging driving dataset for multi-task learning, which possesses geographic, environmental, and weather diversity. For the semantic segmentation task, Bdd100K dataset contains 7000 training images, 1000 validation images and 2000 testing images. The image resolution of Bdd100K is $1280\times720$. 

\begin{table*}[htbp]
 \centering
 \caption{Comparison on Cityscapes val and test set with latest state-of-the-art models. The mIoU and GFlops are calculated using single-scale inference. The ”size” means the image size for GFlops calculation.}
 \vspace{0.1cm} 
  \setlength{\tabcolsep}{4mm}{
\begin{tabular}{@{}ccccccccc@{}}
\toprule
\multirow{2}{*}{Method} &
  \multirow{2}{*}{Publication} &
  \multirow{2}{*}{Backbone} &
  \multicolumn{3}{c}{mIoU (\%)} &
  \multirow{2}{*}{Size} &
  \multirow{2}{*}{ GFLOPs} &
  \multirow{2}{*}{Params(M)} \\ \cmidrule(lr){4-6}
           &             &               & \multicolumn{2}{c}{val}      & test     &           &        &        \\ \midrule
SFNet \cite{SFNet}      & ECCV2020    & ResNe-18      & -            & \multicolumn{2}{c}{78.9} & $2048\times1024$ & 243.87 & 12.87 \\
SegFormer \cite{SegFormer}  & NeurIPS2021 & MiT-B1        & 78.5         & \multicolumn{2}{c}{-}    & $1024\times1024$ & 243.7  & 13.7   \\
NRD \cite{NRD}        & NeurIPS2021 & ResNet-50     & -            & \multicolumn{2}{c}{78.9} & $2048\times1024$ & 234.6  & 36.6  \\
BiAlignNet \cite{BiAlignNet} & ICIP2021    & DFNet2        & 78.7         & \multicolumn{2}{c}{77.1} & $2048\times1024$ & 108.73 & 19.2  \\
 
SFANet \cite{SFANet}     & T-CSVT2022  & ResNet-18     & -            & \multicolumn{2}{c}{78.1} & $2048\times1024$ & 99.6   & 14.6   \\
IFA \cite{IFA}        & ECCV2022    & ResNet-50     & 78.0         & \multicolumn{2}{c}{}     & $1024\times1024$ & 186.9  & 27.8   \\
RTFormer \cite{RTFormer}   & NeurIPS2022 & RTFormer-Base & 79.3         & \multicolumn{2}{c}{-}    & $2048\times1024$ & -      & 16.8   \\
\multirow{2}{*}{DDRNet \cite{DDRNet}} &
  \multirow{2}{*}{TITS2022} &
  DDRNet-23 &
  79.5 &
  \multicolumn{2}{c}{79.4} &
  $2048\times1024$ &
  143.1 &
  20.1  \\
           &             & DDRNet-39      & 80.4         & \multicolumn{2}{c}{80.4} & $2048\times1024$ & 281.2  & 32.3   \\ 
 PIDNet \cite{PIDNet}     & CVPR2023   & PIDNet-M      & 79.9         & \multicolumn{2}{c}{79.8} & $2048\times1024$ & 178.1  & 28.5   \\\midrule
Ours       &             & VAN-S         & {\bf80.8}         & \multicolumn{2}{c}{{\bf80.4}} & $2048\times1024$ & 214.82 & 16.48  \\ \bottomrule
\end{tabular}}
\vspace{-0.3cm} 
\end{table*}
\subsection{Implementation Details}
\label{ssec:Implementation Details}
The experiment is implemented with PyTorch framework. The pretrained Visual Attention Network(VAN) \cite{van} is utilized as the backbone. We use a stochastic gradient decent optimizer with momentum of 0.9 and weight decay of 1e-4. The initial learning rate is 0.01, where the learning rate policy is poly. Random horizontal, random resizing and random cropping are used as data augmentation in the experiment. The scale range of random resizing is [0.5,2.0]. The crop size of random cropping is $1024 \times 1024$ for Cityscapes and $720 \times 720$ for Bdd100K. The batch size is set to 16 for Cityscapes and 32 for Bdd100K. The training epoch is 300 for both two datasets. 

In DNL block, the channel dimension of $q$ and $k$ is reduced to $C/4$, where $C$ is the channel dimension of the input features.
As to the parameters in the loss function, we set $\lambda$ and $\tau$ to be 1 and 0.1 separately.

\begin{table}[]
 \centering
 \caption{Ablation study results on Cityscapes test set. The image size to calculate GFLOPs is $2048\times1024$. PPM and DAPPM  also take concatenated multi-stage features as inputs for fair a comparison. }
 \vspace{0.1cm} 
 \setlength{\tabcolsep}{4mm}{
\begin{tabular}{@{}cccc@{}}
\toprule
Method & mIoU (\%) & GFLOPs & Params(M) \\ \midrule
FRM    & {\bf80.4}     & 214.82   & 16.48   \\
PPM    & 79.7     & 213.21   & 15.23   \\
DAPPM  & 79.1     & 214.81   & 15.98   \\ \bottomrule
\end{tabular}}
 
\end{table}

\subsection{Result Comparison}
\label{ssec:Result Comparison}
 
\noindent  {\bf Experiments on Cityscapes.} Table 1 shows the comparison results with state-of-the-art methods on Cityscapes dataset. The mean of class-wise intersection-over-union (mIoU) is used for accurate comparison. The experimental results demonstrate that our method achieves competitive accuracy with less computation cost. When comparing to the SOTA lightweight network DDRNet-39, the proposed method achieves same 80.4\% mIoU with only 76\% of computational cost and 51\% of params number.

\noindent  {\bf Ablation on Feature Refinement Module design.} We compare the proposed feature refinement module to other context aggregation methods including the pyramid pooling module (PPM) and deep aggregation pyramid pooling module (DAPPM). For a fair comparison, the experiment sets the concatenated multi-stage features as the input of PPM and DAPPM. Table 2 shows the comparison results on Cityscapes test set. Compare with PPM, our feature refinement module increases mIoU by 0.7\% with a small extra calculation burden.

\noindent {\bf Experiments on Bdd100K.} We also evaluate the proposed method on Bdd100K dataset. The results are reported in Table 3. The proposed method achieves 64.9\% mIoU on the Bdd100K val set with 94.55 GFLOPs. There is a 4.3\% increase in mIOU when comparing to SFNet with similar GFLOPs and Params.

\begin{table}[]
 \centering
 \caption{ Comparison on Bdd100K val set. The mIoU and GFlops are calculated using single-scale inference. The image size to calculate GFLOPs is $1280\times720$.}
  \vspace{0.1cm} 
\begin{tabular}{@{}ccccc@{}}
\toprule
Method & Backbone   & mIoU(\%) & GFLOPs & Params  \\ \midrule
HANet \cite{HANet} & ResNet-101 & 64.56    & -        & 65.4M   \\
SFNet \cite{SFNet} & ResNet-18  & 60.6     & 107.34   & 12.87M  \\
PFnet \cite{PFnet}  & ResNet50   & 62.7     & 302.1    & 33.0M   \\ \midrule
Ours   & VAN-S        & {\bf64.9}     & 94.55    & 16.48M  \\ \bottomrule
\end{tabular}
 
\end{table}
 
\section{CONCLUSION }
 
In this paper, we propose a novel semantic segmentation method based on FRM for accurate and fast semantic segmentation. FRM extracts rich semantics by aggregating multi-stage feature maps from the light-weight backbone. By utilizing the DNL block, FRM captures more reasonable global contextual information to further refine the semantic features. The experimental results show that our method achieves a better trade-off between semantic segmentation accuracy and computational cost.

 
%
 
%


\vfill\pagebreak

\small
\bibliographystyle{IEEEbib}
\bibliography{strings,refs}

\end{document}